%% file: main.tex
\documentclass{styles/svproc}

\usepackage[a4paper]{geometry}
\usepackage{url}

\usepackage{amssymb, amsmath, mathtools, physics}
\usepackage{newtxtext}
\usepackage[dvipsnames,svgnames,table]{xcolor}
\usepackage[pdftex,xetex]{graphicx}
\usepackage[shortlabels,inline]{enumitem}
\usepackage{caption}
\usepackage[final,nopatch=footnote]{microtype}

\usepackage[allcolors=Cerulean,colorlinks=true]{hyperref}
\usepackage[square,numbers,compress]{natbib}
\usepackage{siunitx}
\usepackage{setspace,subcaption, wrapfig, nicefrac,booktabs,adjustbox,multirow,pifont,footmisc,authblk,dashrule}
\usepackage[allcolors=Cerulean,colorlinks=true]{hyperref}
\usepackage{multirow}

\usepackage[color=gray!30,textsize=tiny]{todonotes}

\begin{document}

\mainmatter
\title{
EVPeriscope: Extended Perception across Aerial and Ground Vehicles with Event-based Propeller Tracking
}
\titlerunning{EVPeriscope}
\author{Dexter Ong \and Vijay Kumar \and Pratik Chaudhari}
\authorrunning{Dexter Ong et al.}
\institute{General Robotics, Automation, Sensing and Perception (GRASP) Laboratory,\\
University of Pennsylvania, Philadelphia PA 19104, USA}

\maketitle

\begin{abstract}
Reliable relative localization between aerial and ground robots is a key requirement for tightly coordinated heterogeneous teams. This can be difficult to do using conventional frame-based cameras and fiducial markers because they are sensitive to motion blur, lighting variations, and payload constraints.
This paper presents EVPeriscope, an event-based perception system that enables detection, localization and control of a quadrotor using an upward-facing event camera on a ground robot by detecting the high-frequency visual signature of its propellers.
This system allows the quadrotor to function as an extended perception system for the ground robot when onboard sensors exhibit degradation or occlusion.
We demonstrate the capabilities of this marsupial ground-aerial system via experiments in challenging field conditions with wind speeds of up to 15 mph, in both daylight and at night.
We show that the system supports localization and closed-loop navigation through dense foliage where the ground robot’s sensors are occluded.
Our control system for the quadrotor operates at 200 Hz entirely with onboard sensing and computation.
More details and experiment videos can be found on the project page: \url{https://ongdexter.github.io/evperiscope}.
\end{abstract}
%

\input{tex/introduction}
\input{tex/related_work}
\input{tex/method}
\input{tex/experiments}
\input{tex/conclusion}

\bibliography{literature}

\end{document}

%% file: tex/introduction.tex

\section{Introduction}

\begin{figure}
    \centering
    \includegraphics[width=\linewidth]{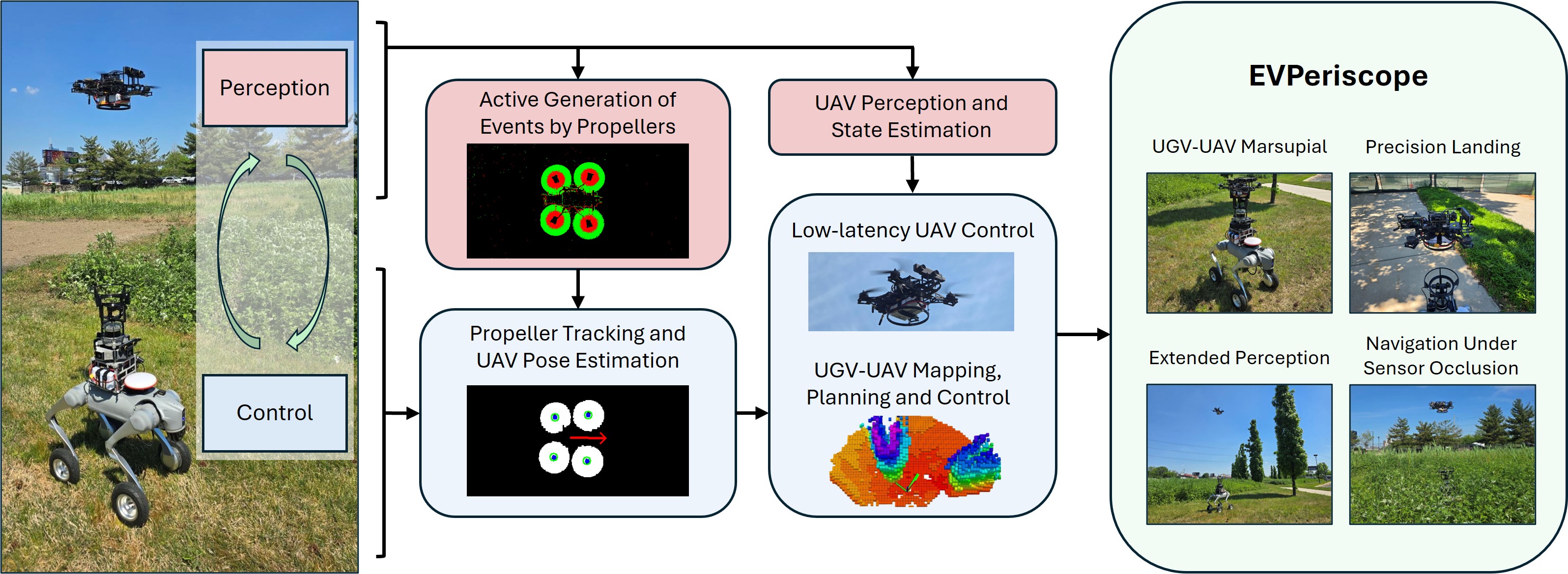}
    \caption{\textbf{EVPeriscope is a marsupial ground-aerial system}.
    It combines the endurance and computational capability of the UGV and the elevated field-of-view of the UAV.
    It uses an upward-facing event camera on the ground robot to capture the high-frequency spatiotemporal structure of the propellers of the quadrotor.
    The propellers are detected, tracked and used for low-latency pose estimation and control of the UAV at 200 Hz.
    This enables precision landing of the UAV on the small footprint of the ground robot which is crucial for long-range deployment of the marsupial system.
    Tightly coupled localization and control between the aerial and ground robot allows the use of the UAV sensors as an extended perception system, enabling behaviors such as mapping and navigation in environments where the UGV sensors are deteriorated or occluded.}
    \label{fig:epa_overview}
\end{figure}

Heterogeneous aerial--ground robot teams are increasingly deployed across a wide range of applications, including search and rescue~\cite{rodriguez2020rescue}, infrastructure inspection~\cite{liao2023cooperative}, and precision agriculture~\cite{ju2020agri}.
These domains require robots that can rapidly adapt their sensing perspective, cover large areas efficiently, and operate reliably under adverse conditions.
Aerial--ground robot teams are particularly well-suited to meeting these demands, as they combine complementary sensing and mobility.
An unmanned ground robot (UGV) can provide endurance, computation, and payload capacity, while an unmanned aerial vehicle (UAV) can rapidly obtain elevated viewpoints over obstacles, clutter, and confined terrain.
Realizing such behaviors, however, requires reliable relative localization between the two robots under fast motion, changing illumination, and onboard computing constraints.

While position estimates provided by a global positioning system (GPS) module may be sufficient if the landing zone is large, vision-based systems can provide higher positioning accuracy for precision landing.
One option is to mount fiducial markers (e.g., AprilTags~\cite{Olson2011AprilTagAR}) on the UAV and track them with a conventional camera.
For small quadrotors, this introduces a practical tradeoff: markers must be large enough to detect reliably, yet they add weight, occupy limited surface area, and may interfere with airflow around the vehicle~\cite{sanket2021evpropnet}.
A fiducial marker could be placed on the landing zone~\cite{lee2024fiducial,springer2022gimbal} but this requires the quadrotor to have an additional downward-facing camera just for takeoff and landing.
In addition, the same size constraints on the size of the fiducial marker also apply to smaller landing zones such as small ground robots.
These limitations motivate a markerless approach that uses visual structure already present on the aerial vehicle.

Event cameras are well-suited to this setting because they can capture rapid brightness changes at a microsecond temporal resolution and have a high dynamic range which enables robust operation under extreme lighting changes.
In contrast, even high-speed frame-based cameras suffer from motion blur, limited dynamic range, and degraded performance in low-light or high-contrast conditions.
Spinning propellers produce strong, localized event activity that remains detectable even when conventional imaging fails.
We leverage this to estimate propeller locations and, using their known geometry, recover the quadrotor pose for relative localization and control between the ground and aerial robot.
This enables extended perception: the UAV acts as a periscope to augment the UGV's sensing.

An overview of the EVPeriscope system is provided in Fig.~\ref{fig:epa_overview}.
This paper presents a lightweight event-based pipeline for quadrotor localization and control using an upward-facing event camera on a ground robot.
This enables the use of the quadrotor as an extension of the ground robot's perception system to augment its sensing capabilities.
The contributions of this paper are as follows.
\begin{enumerate}
    \item
    A real-time markerless event-based relative localization system enabling closed-loop UAV tracking from an upward-facing UGV-mounted event camera at 200 Hz under challenging outdoor conditions.
    \item
    A ground-aerial marsupial system with a tightly coupled extended perception architecture in which the UAV functions as an extendable sensor for the UGV like a periscope.
    \item
    Field validation demonstrating robust operation in dense foliage, nighttime conditions, and wind disturbances using fully onboard sensing and computation.
\end{enumerate}

%% file: tex/related_work.tex

\section{Related Work}

\subsection{Event-based UAV Detection}

Event-based propeller detection leverages neuromorphic cameras to exploit the fast, periodic motion of UAV propellers, enabling robust detection and tracking of UAVs under challenging lighting and background clutter. EVPropNet~\cite{sanket2021evpropnet} models propeller geometry to generate synthetic event streams; event data is converted into an image for propeller detection by a convolutional neural network and used for UAV detection, tracking and landing.
Spiking neural networks have also been used for detection of UAVs in low-power applications~\cite{eldeborg2024drone}.
More recent pipelines like EventPro~\cite{chen2025count,spetlik2025efficientpropeller} focus on dynamics by estimating propeller rotational speed via motion-compensated event batches.

With the proliferation of accessible and low-cost consumer UAVs, UAV detection is a crucial capability for autonomy in robotics, but also for safety and security.
Multiple datasets have been created to help drive research in event-based UAV detection~\cite{pawelczyk2020realworld,chen2025event,magrini2025drone}.
Event-based detection of UAVs has also been proposed for the application of virtual fences in surveillance~\cite{terrence2022dvxplorer}.
We use a simple and lightweight approach for detection and tracking of propellers from a ground robot to enable low-latency localization and control of the quadrotor.
In contrast to learning-based detectors, this representation operates directly on the incoming events and requires neither a trained model nor propeller-specific training data.
We present an application of event-based UAV detection for persistent relative localization of the full pose of the UAV for tightly coupled heterogeneous coordination.

\subsection{UAV Landing Systems}
GPS can provide position information to aid autonomous landing of UAVs~\cite{wang2015gpslanding}.
However, without RTK correction which is often unavailable in the field, GPS can have significant noise, on the order of meters, which is inadequate for precision landing.
Ultra-wideband positioning systems have been used to achieve positioning accuracy on the order of tens of centimeters~\cite{xia2022uwblanding}.
Even so, when the landing zone is small due to space constraints, such as on a small ground robot, this accuracy is insufficient.
Vision-based landing systems~\cite{araar2017visionlanding,zhao2022landing,polvara2018landing} localize the landing pad using fiducial markers which can provide more accurate 6-degree-of-freedom (6-DoF) pose estimates when the marker is clearly visible.
However, these methods are constrained by the size and capacity of both the ground and aerial robots, requiring a marker or downward-facing camera mounted on the UAV.
They also suffer from limitations of conventional frame-based sensors, resulting in degradation of detection performance under adverse illumination conditions and motion blur~\cite{raxit2024yolotag,springer2022autonomous}.
We exploit neuromorphic perception to perform robust, target-less and low-latency quadrotor detection and tracking, leveraging the fact that quadrotor propellers actively generate a distinctive high-frequency temporal signature in the event stream.

\subsection{Marsupial Ground-Aerial Systems}
Marsupial robot teams use a carrier robot to transport, deploy, recover, or support one or more child robots, combining the endurance and payload capacity of a larger platform with the mobility and viewpoint diversity of smaller agents~\cite{murphy1999marsupial}.
Most approaches utilize the UAV to achieve a complementary objective with the UGV for exploration and mapping~\cite{hudson2022heterogeneous,de2022marsupial,zacharia2025collaborative}.
These approaches typically deploy the UAV from the UGV once per mission and do not account for recovery and multiple deployments.
Instead of treating the marsupial system as independent robots after the initial deployment, this work explores the use of the UAV as a tightly integrated extension of the UGV perception system where the UAV is dynamically deployed and recovered.

%% file: tex/method.tex

\section{Method}

\begin{figure}
    \centering
    \includegraphics[width=0.95\linewidth]{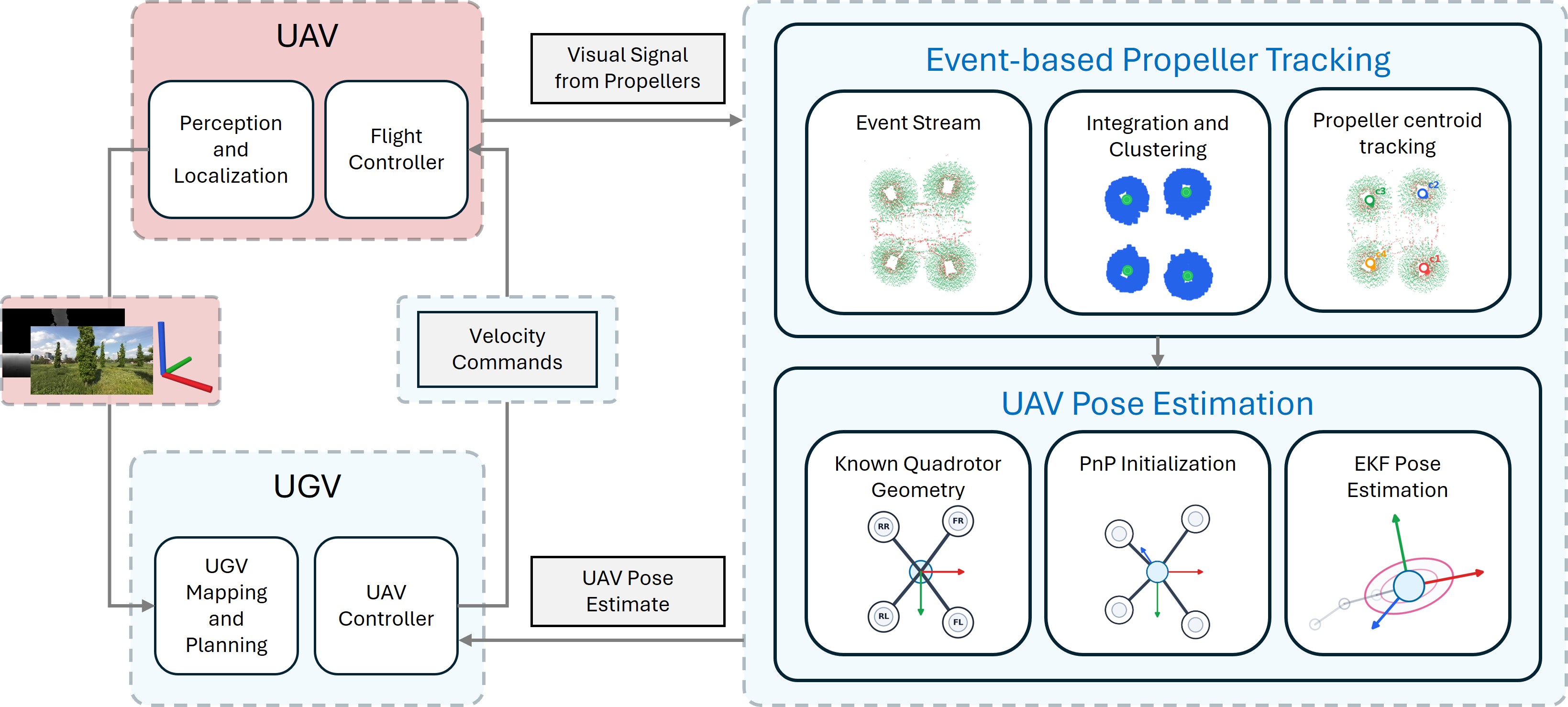}
    \caption{\textbf{EVPeriscope system diagram.} The UGV-mounted event camera detects and tracks the UAV propellers from asynchronous events, clusters them into propeller centroids, and uses the known asymmetric quadrotor geometry to estimate the UAV pose. The estimated pose is then used by the UGV controller to command the UAV through velocity commands. UAV observations and state estimates are transmitted to the UGV as extended perception that can be used for UGV autonomy.}
    \label{fig:system_diagram}
\end{figure}

The EVPeriscope system consists of an event-based propeller detection and tracking pipeline, quadrotor pose estimation module and UAV controller.
The system diagram is shown in Fig.~\ref{fig:system_diagram}.

\subsection{Event-based Propeller Detection and Tracking}

Event cameras produce sparse, high-temporal-resolution measurements that are inherently asynchronous and noisy, making instantaneous frame-based detection unreliable.
To robustly capture the spatiotemporal structure of rapidly rotating propellers against noise and background events generated by the environment, we maintain a continuously updated state that emphasizes recent events while suppressing older ones.

The propeller detector maintains a persistent state ``image'' \(S(u,v)\) with $u, v \in \mathbb{Z}$ that is updated once per event batch using a first-order exponential IIR filter on the event stream to result in leaky temporal integration.
At the end of each batch, the previous state is decayed by a global factor.
Each event contributes to the pixel where it occurred with a weight that depends on how recently it occurred. Altogether,
\begin{equation}
    S(u,v) \leftarrow S(u,v)\exp(-\frac{(t_{\text{end}}-t_{\text{last}})}{\tau})
    + \sum_{i=1}^{n} \exp(-\frac{(t_{\text{end}}-t_i)}{\tau})\delta_{u, u_i}\delta_{v, v_i}.
\end{equation}
where \(S(u,v)\) denotes the filter state at pixel \((u,v)\), the exponential time constant is \(\tau\), the end time of the current batch is \(t_{\text{end}}\), end time of the previous batch is \(t_{\text{last}}\), the number of events in the batch is \(n\), and \(t_i\) is the timestamp of event \(i\).
The Kronecker delta function is non-zero only when $u = u_i$ and $v = v_i$ for the event pixel \((u_i,v_i)\).
To maximize response to the fast-moving propellers, all events are treated as positive contributions regardless of their polarity.
We process the event stream in \( 5 \) ms windows, i.e., $t_{\text{end}}-t_{\text{last}} = 5$ ms, and decay the representation with \(\tau = 20 \) ms.
These parameters were fixed across all experiments.

Pixels in $S$ with intensity one standard deviation above the mean (both computed only on pixels with non-zero intensities) are used to obtain a binarized image. This image is cleaned using morphological erosion, followed by dilation, to suppress background noise.
The $k$-means algorithm is used to obtain candidate centroids on this cleaned imaged.
With $k=4$, we obtain clusters corresponding to events from each of the propellers.
We initialize the $k$-means algorithm using centroids detected in the previous image.
This technique can, in principle, be extended to an $N$-rotor platform by setting $k=N$.
The persistent state ensures that our detector for the centroids is stable over time and clusters depict clear structure in the propellers.
The tracker maintains temporal continuity by initializing the estimates of the centroids from the previous frame and refining them using the regions that are detected in the current frame.

\subsection{Quadrotor Pose Estimation}

We assume that the quadrotor geometry is known and that the frame is asymmetric, with the front arms differing in length and relative angle from the rear arms. This asymmetry allows us to use the centroids of the propellers detected in the previous section to estimate the full 6-DoF pose of the quadrotor.
If the quadrotor is symmetric or its geometry is unknown, the position of the quadrotor can still be estimated for tracking and landing, similar to what is done in prior work~\cite{sanket2021evpropnet}.
However, yaw would not be observable in such cases.
Geometric correspondence is established in our case by ordering the centroids.
Let the centroid of the $i^{\text{th}}$ propeller in the image plane be
$\mathbf{z}_i \in \mathbb{R}^2$ and denote
\[
  \mathbf{z} =
  [\mathbf{z}_1^\top,\mathbf{z}_2^\top,
   \mathbf{z}_3^\top,\mathbf{z}_4^\top]^\top.
\]
Let the mean be
$\bar{\mathbf{z}} = \frac{1}{4}\sum_i \mathbf{z}_i$
and their orientation with respect to the mean be
\[
\gamma_i =
\tan^{-1}\left(
\frac{(\mathbf{z}_i-\bar{\mathbf{z}})_y}
     {(\mathbf{z}_i-\bar{\mathbf{z}})_x}
\right).
\]
For our platform, the front pair of propellers has the largest distance between them. In this work, we require visibility of all four propeller centroids, although in principle three propellers would be sufficient.
Since we know the propeller locations in the quadrotor body frame exactly, an initial pose is obtained using Perspective-n-Point (PnP).
This estimate is updated with a discrete-time Extended Kalman Filter (EKF). The state is
\[
\mathbf{x}
\equiv
[\mathbf{p}^\top,
 \boldsymbol{\Theta}^\top,
 \mathbf{v}^\top,
 \boldsymbol{\omega}^\top]^\top
\in \mathbb{R}^{12},
\]
where $\mathbf{p} \in \mathbb{R}^3$ is the quadrotor position
in the camera frame,
$\boldsymbol{\Theta}=[\phi,\theta,\psi]^\top$ is the attitude,
and $\mathbf{v},\boldsymbol{\omega}\in\mathbb{R}^3$ are the
translational and angular velocities.
We assume constant velocity dynamics.
Observations are given by the propeller centroids
$\mathbf{z} \in \mathbb{R}^8$ in the image plane.
For the $i^{\text{th}}$ propeller, let
\[
\mathbf{q}_i =
\begin{bmatrix}
X_i & Y_i & Z_i
\end{bmatrix}^{\top}
\in \mathbb{R}^3
\]
denote its position in the camera frame. The corresponding image measurement is
\[
\mathbf{z}_i = \mathbf{h}_i(\mathbf{x}) \equiv
\begin{bmatrix}
c_x + f_x \dfrac{X_i}{Z_i} \\
c_y + f_y \dfrac{Y_i}{Z_i}
\end{bmatrix}.
\]
Here, $\mathbf{q}_i$ depends on the quadrotor attitude
$\boldsymbol{\Theta}$, position $\mathbf{p}$, and known propeller geometry.
The camera focal lengths are $(f_x,f_y)$ in pixels, and
$(c_x,c_y)$ denotes the principal point.

\subsection{UAV Control}

Given the estimated UAV pose from the EKF, we generate UAV body-frame velocity commands as follows.
The event camera is aligned with the body frame of the ground robot.
In this work, we set the target state for the quadrotor to be centered on the camera with the yaw aligned with the front of the ground robot.
We therefore design a UAV velocity controller to track the error in position $\mathbf{p}$ and yaw $\psi$ in the camera frame of the ground robot.
For each axis, a standard PID controller determines the control,
with separate gain sets for planar motion, vertical motion, and yaw.
This velocity is transformed into the frame using the estimated yaw in the EKF before publishing to the UAV.
This ensures that the controller output is consistent with the vehicle command interface while still regulating position in the camera reference frame.

\subsection{UAV Takeoff and Landing}

The UAV is first commanded to take off to a set point of 1.5 m above the ground robot.
EVPeriscope is then initialized on the UGV to start the propeller detection and tracking modules.
Upon stable localization of the UAV, the EVPeriscope controller begins streaming velocity commands to the UAV.
The quadrotor is then transitioned to a tracking state and control authority is handed off to the UGV.

When the landing sequence is triggered, the UAV is commanded to a height of 0.5 m.
To ensure landing under stable wind conditions, the controller checks that the UAV is within 0.05 m of the target setpoint for at least 3 s before proceeding.
When this condition is satisfied, the PX4 land mode is triggered. 

\subsection{Platform Details}

\begin{figure}[tb]
    \centering
    \includegraphics[width=0.98\linewidth]{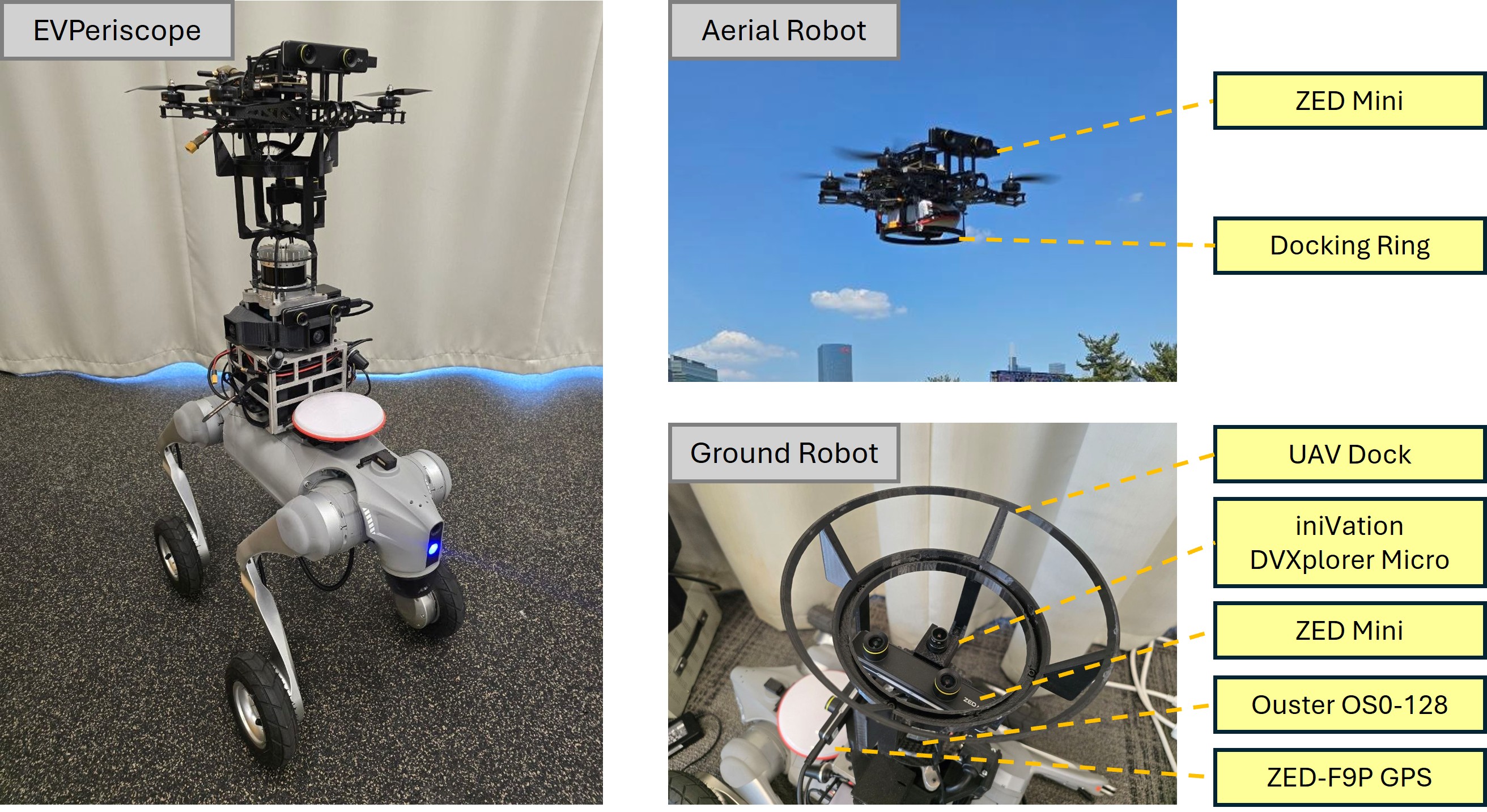}
    \caption{\textbf{System hardware configuration.} EVPeriscope (left) consists of a custom-built quadrotor (upper right), and a Unitree Go2-W with a custom sensor stack  (lower right). The core sensors of the system include the ZED Mini on the UAV for perception and state estimation, and the iniVation DVXplorer Micro event camera on the UGV for propeller tracking. A ZED Mini, Ouster OS0-128 and ZED-F9P GPS module on the UGV are also used for baseline comparisons. A custom docking system supports carrying and landing the UAV.}
    \label{fig:system_overview}
\end{figure}

An overview of the platform is shown in Fig.~\ref{fig:system_overview}.
The Kestrel UAV is a custom-built quadrotor with a 31.2 cm wheelbase and 7-inch propellers.
The platform has a PX4 flight controller and is equipped with a forward-facing Stereolabs ZED Mini for state estimation.
The sensors on the Unitree Go2-W UGV include an iniVation DVXplorer Micro event camera, upward-facing and forward-facing ZED Mini cameras, an Ouster OS0-128 LiDAR and a ZED-F9P GPS module.
The event camera has a resolution of 640 $\times$ 480 and an FOV of approximately 90° $\times$ 70°.
The UGV has an NVIDIA Jetson AGX Orin for compute.
Communication between robots is supported by Rajant mesh radios.
A custom-built docking system supports carrying and landing the UAV.
The docking ring on the UAV is 14 cm in diameter.
The docking pad on the UGV has a matching ring and is 30 cm in diameter at the other end.

The software stack runs on ROS 2 Humble over Zenoh middleware. 
All event processing, propeller tracking, PnP/EKF pose estimation, and relative-pose control are executed on the UGV's Jetson AGX Orin.
The UAV computes depth and visual-inertial odometry onboard using the ZED Mini and Orin NX.
The UAV transmits these perception and state-estimation outputs to the UGV, while the UGV sends velocity commands to the UAV flight controller.

The ground robot runs the UAV detection and control modules on the Orin AGX at a fixed rate of 200 Hz --- this includes communication with the UAV.
The event stream is processed in 5 ms non-overlapping windows.
Each iteration of our control loop takes on average 2.53 ms, with all the modules---propeller detection and tracking, UAV pose estimation and the controller---running synchronously.

%% file: tex/experiments.tex

\section{Experiments}

\begin{table}[tb]
    \centering
    \caption{Quadrotor pose estimation error at different altitudes above the UGV. The position and attitude estimates from EVPeriscope are compared against the estimate state from the UAV onboard VIO. The RMSE and standard deviations of position and attitude errors over each altitude sequence are reported.}
    \label{tab:pose_error}
    \setlength{\tabcolsep}{8pt}
    \begin{adjustbox}{width=\linewidth}
    \renewcommand{\arraystretch}{1.2}
    \begin{tabular}{c ccc ccc}
        \toprule
        \multirow{2}{*}{\textbf{Altitude}} & \multicolumn{6}{c}{\textbf{Position (m ) / Attitude (deg) Error}} \\
        & X & Y & Z & Roll & Pitch & Yaw \\
        \midrule
        0.5 & 0.014 {\tiny$\pm$0.006} & 0.027 {\tiny$\pm$0.007}  & 0.017 {\tiny$\pm$0.005} & 1.975 {\tiny$\pm$0.915} & 1.894 {\tiny$\pm$1.137} & 1.342{\tiny$\pm$0.103} \\
        1.0 & 0.065 {\tiny$\pm$0.029} & 0.086 {\tiny$\pm$0.015} & 0.023 {\tiny$\pm$0.014} & 2.250 {\tiny$\pm$1.856}  & 1.970 {\tiny$\pm$1.522} & 1.611 {\tiny$\pm$0.155} \\
        2.0 & 0.084 {\tiny$\pm$0.015}  & 0.125 {\tiny$\pm$0.024} & 0.129 {\tiny$\pm$0.019} & 2.744 {\tiny$\pm$2.419} & 5.162 {\tiny$\pm$3.645} & 1.944 {\tiny$\pm$0.320} \\
        4.0 & 0.142 {\tiny$\pm$0.040} & 0.434 {\tiny$\pm$0.035} & 0.493 {\tiny$\pm$0.032} & 4.248 {\tiny$\pm$3.131} & 9.432 {\tiny$\pm$9.344} & 4.752 {\tiny$\pm$0.576} \\
        6.0 & 0.364 {\tiny$\pm$0.084} & 0.546 {\tiny$\pm$0.051} & 0.665 {\tiny$\pm$0.055} & 7.260 {\tiny$\pm$6.914} & 8.760 {\tiny$\pm$6.537} & 8.500 {\tiny$\pm$0.868} \\
        \bottomrule
    \end{tabular}
    \end{adjustbox}
\end{table}

\subsection{UAV Detection and Pose Estimation}
We first compute the error of the estimated quadrotor pose at different distances from the event camera to understand the performance of our UAV detection and localization method.

\subsubsection{Setup}
The ground robot is stationary and the UAV is commanded to a setpoint at different heights above the UGV.
1 minute of data is collected per sequence at various heights up to 6 m.
We use the pose estimate from ZED VIO as the reference estimate of the UAV pose.

We also compare our approach to fiducial-based detection methods with conventional frame-based cameras and identify failure cases of such methods.
We consider two scenarios: an RGB camera on the ground robot facing upwards to detect a fiducial marker mounted on the quadrotor, and a downward-facing RGB camera mounted on the quadrotor detecting a fiducial marker on the ground robot.
We use a 0.1 m $\times$ 0.1 m AprilTag marker that matches the footprint of the UAV docking ring.

Finally, we conduct an experiment to demonstrate operation in extreme low-light conditions.
The same concept of active event generation with propellers is used in the dark by mounting LED propellers on the UAV.
Since VIO is severely degraded in such illumination conditions, GPS is used for state estimation in the \textit{night} experiment.
This is the only experiment using GPS.

\subsubsection{Results}

\begin{figure}[!tb]
    \centering
    \includegraphics[width=0.98\linewidth]{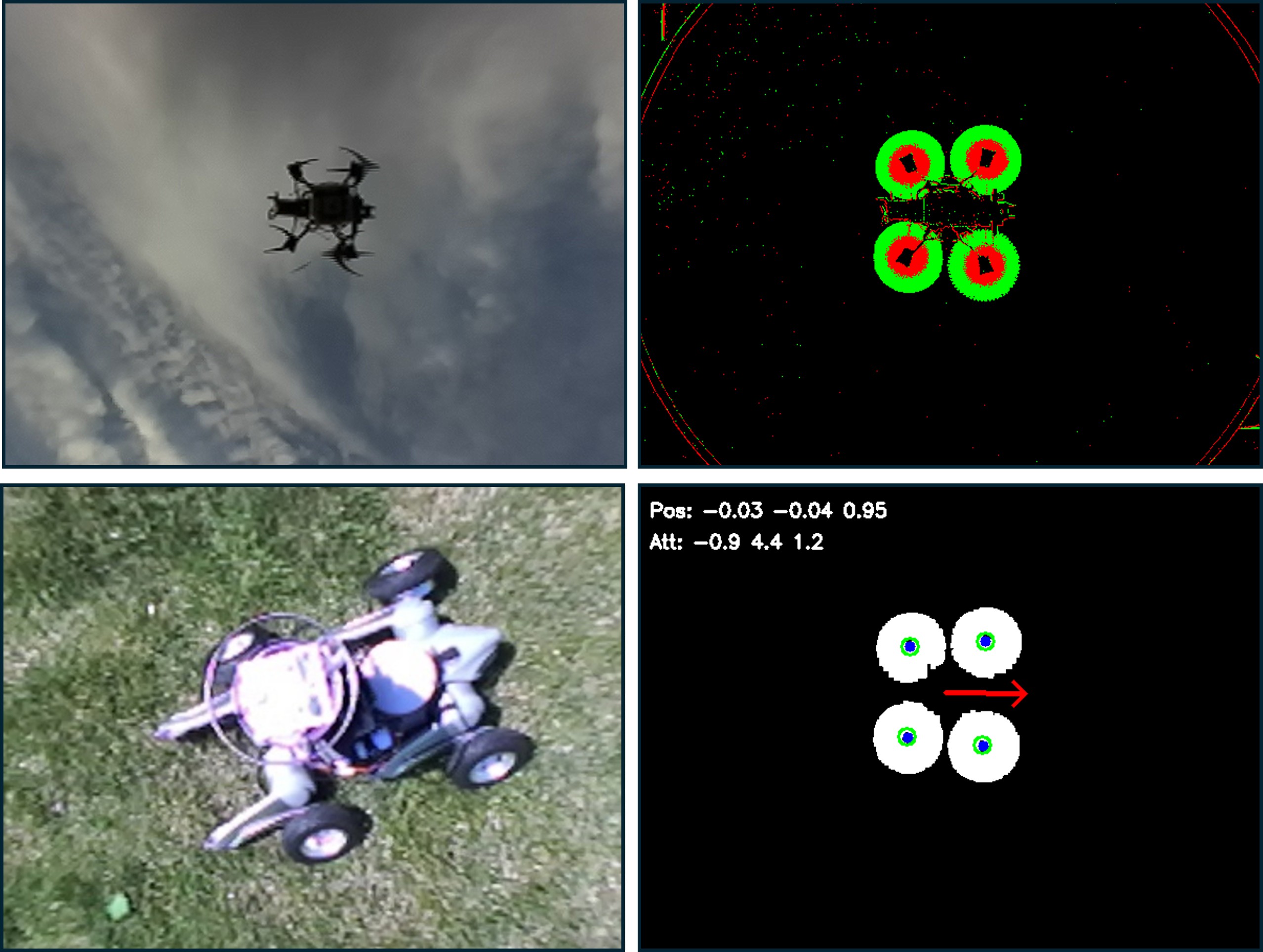}
    \caption{\textbf{Failure cases of UAV localization with conventional cameras.} Even in overcast conditions, upward-facing imagery is often underexposed due to the bright backdrop of the sky (upper left). On the other hand, downward-facing cameras may struggle to handle the strong reflections of the fiducial markers (lower left). Conventional cameras likewise fail in low illumination conditions (middle). The harsh lighting conditions of outdoor environments make it challenging for fiducial marker detection using conventional frame-based cameras. In contrast, event-based cameras provide a strong signal on the propellers across a wide range of lighting conditions (upper right). The blue circles denote the estimated 2D propeller centroids, the green circles are the projected 3D centroids, and the red arrow indicates the estimated yaw of the UAV (lower right).}
    \label{fig:det_day}
\end{figure}

\begin{figure}[tb]
    \centering
    \includegraphics[width=0.98\linewidth]{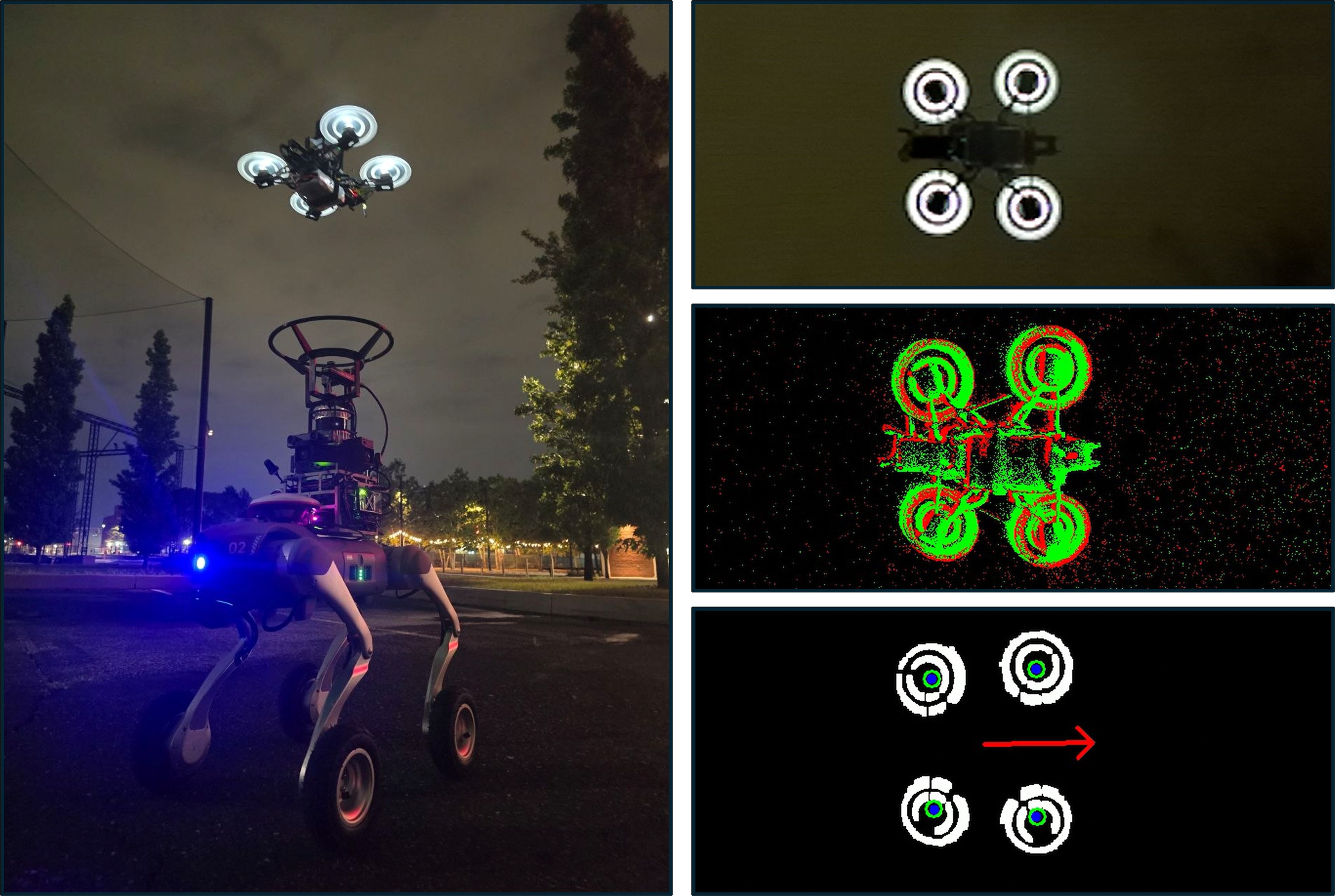}
    \caption{\textbf{Nighttime operation with EVPeriscope.} The ground robot with the deployed aerial robot is shown on the left. The RGB camera image (upper right), event stream (middle right) and tracked propellers and estimated UAV pose (lower right) are visualized. With LED propellers to aid event generation, the framework is capable of propeller detection and tracking in low-light conditions.}
    \label{fig:det_night}
\end{figure}

Table~\ref{tab:pose_error} summarizes the quadrotor pose estimation error at different altitudes.
We present the Root Mean Square Error (RMSE) of the pose estimate for positions and attitude along different axes along with their standard deviations.
As expected, the farther the UAV from the camera, the higher the error in the pose estimation due to the resolution and field-of-view of the event camera.
We show that we can localize the UAV with a reasonable margin of error for control---the positional error is less than 0.03 m when the quadrotor is near the camera and less than 0.7 m error at 6 m height.
At larger distances, pose estimation accuracy is limited by the resolution and FOV of the event camera.
In this work, we use a lens with a wider FOV to enable good tracking at small distances so as to support precise autonomous landing.
Depending on the application, a narrower FOV may be preferred to improve tracking of the UAV at higher altitudes.

We highlight the challenges of frame-based detection of fiducial markers in Fig.~\ref{fig:det_day}.
Conventional cameras lack sufficient dynamic range to balance exposure due to the sunlight, and struggle in darker conditions and farther distances as motion blur becomes more prominent.
We ran the same set of experiments with the ZED Mini and a 100 mm $\times$ 100 mm AprilTag marker and found that marker detections were inconsistent at 0.5 m from the camera and failed beyond that.
This is due to the fact that the platform size limits the size of the markers and harsh outdoor lighting conditions also lead to deterioration.
Our approach can consistently detect the propellers and quadrotor across lighting conditions.
This underscores the advantages of event-based cameras when operating out in the field under challenging conditions such as harsh sunlight.

Lastly, we show qualitative results of EVPeriscope operating at night in Fig.~\ref{fig:det_night}.
With LED propellers, the aerial platform can still generate a strong signal in the event stream, enabling detection and tracking of the UAV with our method even in the dark.
This demonstrates the potential of EVPeriscope for operation throughout the day and night regardless of lighting conditions.

\subsection{Localization under Sensor Occlusion}

\begin{figure}[tb]
    \centering
    \includegraphics[width=0.99\linewidth]{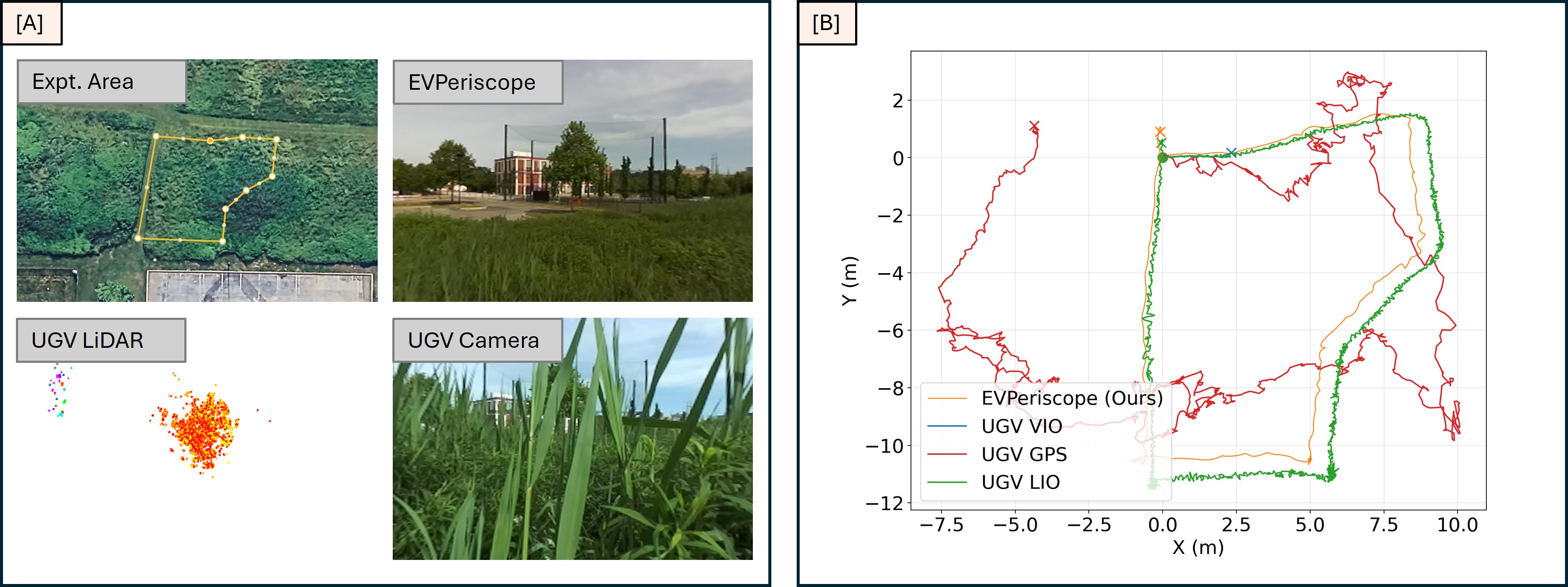}
    \caption{\textbf{Visualization of results for localization in dense foliage.} [A] The ground robot is teleoperated to perform a loop through the field (top left). Sensor measurements of the periscope and ground robot are visualized. The UGV sensors are occluded by the tall grass whereas the UAV is unobstructed. [B] The estimated trajectories from each sensor are visualized. The start of each trajectory is denoted by $\circ$ and the end by $\times$. EVPeriscope is able to maintain a high vantage point above the foliage to enable accurate localization.}
    \label{fig:epa_loc}
\end{figure}

We next show that our approach enables a ground robot to localize itself within dense foliage.
By leveraging the extended perception provided by the quadrotor in the EVPeriscope framework, we show that the UGV is able to maintain accurate localization for navigation even when its own navigation sensors are occluded.

\subsubsection{Setup}
We manually tele-operate the UGV through dense foliage in a field.
We record the VIO state estimate from the quadrotor in EVPeriscope, the VIO estimate from the forward facing RGB camera on the UGV, the LiDAR-inertial odometry~\cite{malladi2026rko} (LIO) from the LiDAR on the UGV, and the GPS measurements of the UGV for ground truth.

\subsubsection{Results}
The results are presented in Fig.~\ref{fig:epa_loc}.
The forward-facing ZED VIO completely fails due to occlusion of the sensor.
Since the GPS antenna is not mounted high enough on the robot to be above the foliage, GPS measurements also exhibit a large amount of noise, highlighting the importance of alternative localization methods in such scenarios.
Due to the wide field-of-view, high scan density and long range of the LiDAR sensor, the LIO is still able to capture global structure from faraway trees and lampposts and maintains a reasonable state estimate overall.
However, there is significant noise along the trajectory which is detrimental to mapping and navigation.
This is also dependent on how high the LiDAR can be mounted on the ground robot.
Additionally, a majority of the LiDAR points are still occluded as shown in Fig.~\ref{fig:epa_loc} and building a detailed map of the field with the LiDAR would not be possible.
EVPeriscope maintains good perception and localization above the foliage for navigation.

\subsection{Mapping with Extended Perception}

\begin{figure}[tb]
    \centering
    \begin{minipage}{0.48\linewidth}
        \centering
        \includegraphics[width=\linewidth]{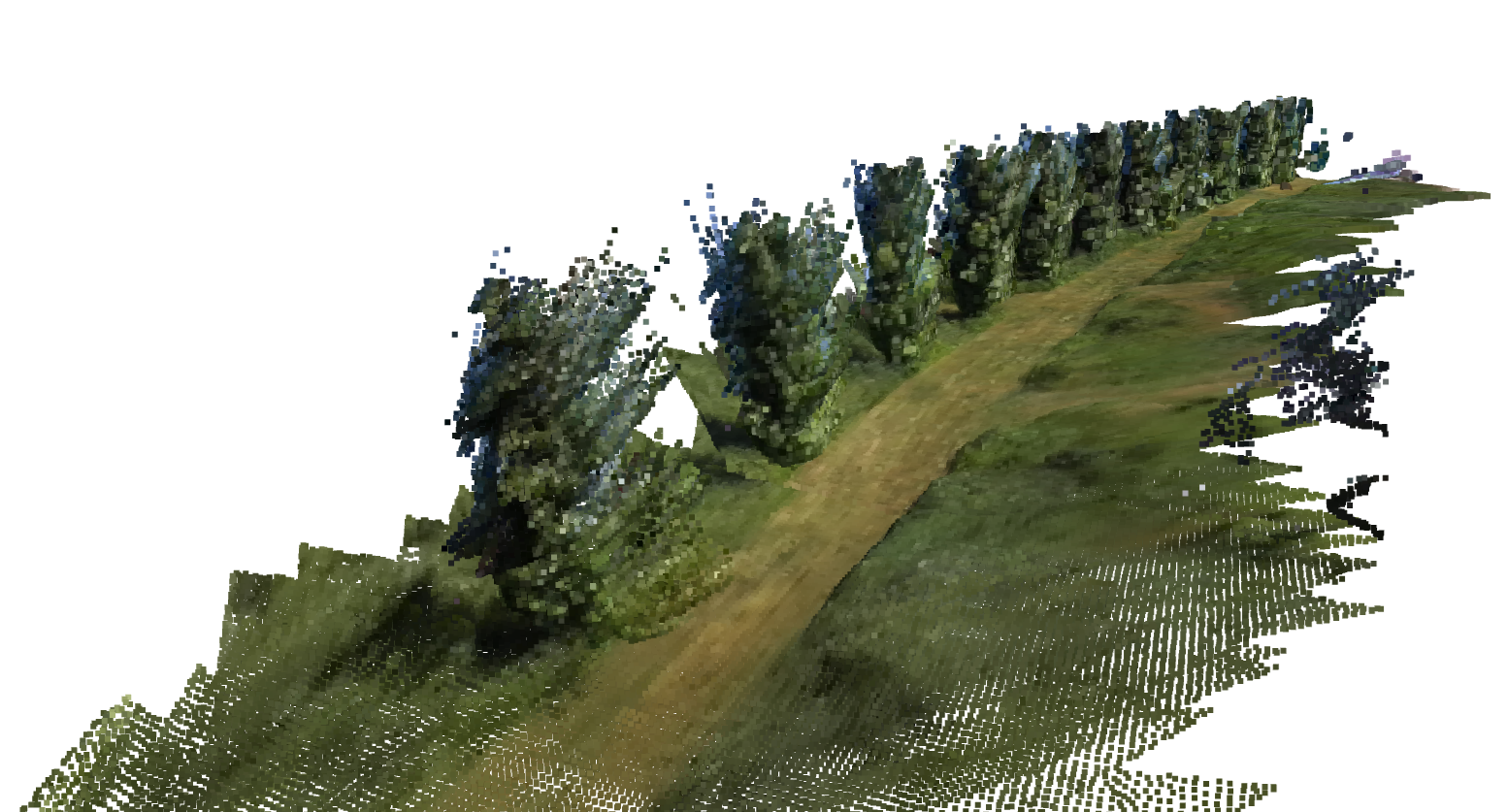}
    \end{minipage}%
    \hspace*{2ex}
    \begin{minipage}{0.425\linewidth}
        \centering
        \resizebox{\linewidth}{!}
        {
        \begin{tabular}{l r}
            \toprule
            \textbf{Method} & \textbf{Volume Mapped (m$^3$)} \\
            \midrule
            UGV LiDAR & 421.82 \\
            UGV Camera & 146.13 \\
            EVPeriscope & 314.25 \\
            \bottomrule
        \end{tabular}
        }
    \end{minipage}
    
    \caption{
        \textbf{Visualization of the map constructed by EVPeriscope (left) and the volume mapped by each method (right).}
        The UGV camera is limited in its field‑of‑view due to being close to the ground.
        The UGV LiDAR, with its larger range and field‑of‑view, is provided as an upper bound.
        EVPeriscope is able to map more of the environment compared to the UGV camera with the elevation advantage of the UAV. This is crucial when the areas of interest are at a much higher elevation than the ground robot, such as when mapping trees in orchards or infrastructure inspection.
    }
    \label{fig:uav_map}
\end{figure}

We investigate the capability of EVPeriscope to enable mapping of areas of the environment that are typically beyond the field-of-view of UGV sensors by leveraging the elevated perspective of the UAV periscope.

\subsubsection{Setup}
The system is tele-operated along a row of trees in the field with the objective of constructing a map of the environment.
We build a map from the measurements from the UGV front-facing camera, the EVPeriscope UAV camera and the LiDAR on the UGV.
The map constructed by each method is voxelized to 0.1 m and the occupied volume of each map is computed.

\subsubsection{Results}
Results are presented in Fig.~\ref{fig:uav_map}. EVPeriscope is able to map more of the environment compared to the UGV camera.
This is particularly useful when parts of the environment are beyond the vertical field-of-view of the UGV sensors, such as buildings and trees.
EVPeriscope can leverage the elevation advantage provided by the UAV to map these areas of interest.
This provides the benefits of a UAV for mapping while enabling longer missions by having the UGV carry the UAV between flights.

\subsection{Closed-loop Navigation in Dense Foliage}
We demonstrate the full EVPeriscope system with closed-loop navigation and obstacle avoidance in dense foliage.

\subsubsection{Setup}
The experiment setup is visualized in Fig.~\ref{fig:epa_obs}.
The system is given a waypoint behind the tree.
To reach the waypoint, the system has to navigate through tall grass while avoiding the tree.
The UAV is first tasked to take off from its dock on the UGV.
Once the UAV is detected in the event stream, it is then transitioned to the tracking mode and positioned at 1.5m above the ground robot.
Sensor measurements including RGB and depth images, and state estimation from the periscope are transmitted to the UGV in real time which are then fused into a map for planning and obstacle avoidance.
Finally, navigation to the waypoint is initiated.

Depth images and corresponding poses are integrated into an occupancy map.
For this experiment, we approximate the UAV-to-UGV transform as fixed at the commanded tracking setpoint. The relative pose estimated by EVPeriscope can instead be used to update this transform online.
A global search with \(A^*\) on the occupancy map provides a coarse obstacle-free path, while a motion primitive-based planner is used to obtain dynamically feasible and collision-free trajectories (shown in green in Fig.~\ref{fig:epa_obs}).
A trajectory tracker computes velocity commands which are sent to the robot control interface.

We also evaluate the tracking accuracy of the periscope while the system is navigating the environment.
The experiment was conducted in wind conditions of 10--15 mph.

\subsubsection{Results}
\begin{figure}[tb]
    \centering
    \includegraphics[width=\linewidth]{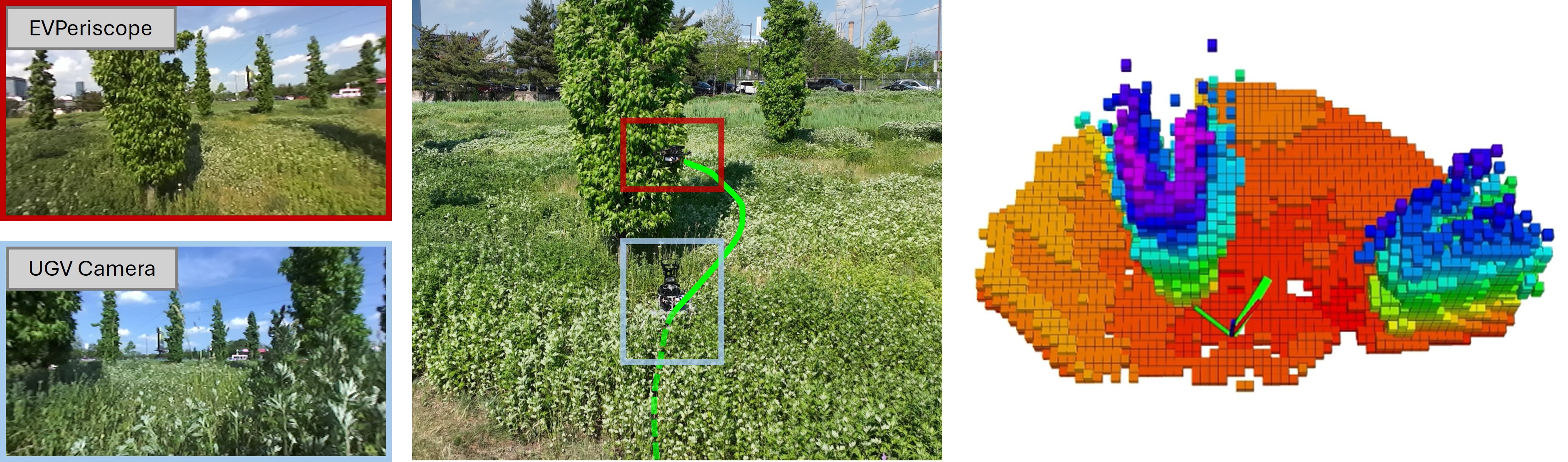}
    \caption{\textbf{EVPeriscope enables closed-loop navigation and obstacle avoidance in dense foliage where the UGV sensors are occluded.} The sensor measurements of each robot are shown (left). The robots are highlighted in the scene (middle). Depth images and state estimates of the UAV are transmitted to the UGV, which performs mapping, planning and trajectory tracking with its onboard compute (right). The robot is indicated by the colored axes. The planned trajectory to avoid the obstacles is visualized as a green path.}
    \label{fig:epa_obs}
\end{figure}

Qualitative results of the demonstration are visualized in Fig.~\ref{fig:epa_obs}.
The system enables the robots to cooperatively navigate through the dense foliage while avoiding the tree.
Along the trajectory, the closed-loop tracking RMSE is 0.125 m on position and 0.916$^\circ$ on yaw.
This experiment demonstrates the capabilities of the EVPeriscope system, from robust quadrotor detection and tracking, to sharing of sensor measurements across robots for collaborative autonomy, with all modules running in real time on the platforms with onboard sensors and compute.

%% file: tex/conclusion.tex

\section{Discussion and Limitations}

When the UGV sensors are occluded, the quality of the state estimation of the EVPeriscope system is limited by the localization sensor and algorithm on the UAV---visual-inertial odometry in this case.
A larger UAV could be equipped with a LiDAR sensor for much more accurate localization, though payload constraints of the UGV have to be considered.
While this work can filter out events generated by the quadrotor frame and background noise, it cannot handle high frequency noise such as from the propellers of another UAV.
Future work could explore identifying and assigning propellers to unique UAVs to enable tracking and control of multiple robots.

When the ground robot's sensors are unobstructed, the localization of both the UAV and UGV can be more tightly coupled, particularly since the UGV benefits from higher accuracy due to its LiDAR system.
Although the UAV VIO-based state estimate may accumulate drift over time, the relative pose between the two platforms remains reliable.
By fusing this relative pose information with the more accurate UGV state estimation, the system can leverage the aerial robot's perception capabilities while minimizing the effects of its localization drift, making the most of both modalities.

\section{Conclusion}

We presented EVPeriscope, a marsupial UGV--UAV system that uses an upward-facing event camera on a ground robot to detect, localize, and control a quadrotor.
The system enables tightly coupled coordination without external infrastructure.
Through real-world experiments, we showed that event-based propeller detection remains reliable under challenging outdoor lighting, supports localization and navigation when ground-level sensors are occluded by dense foliage, and expands mapping coverage by using the UAV as an elevated sensing platform.
These results demonstrate the potential of this approach for robust heterogeneous robot teams in GPS-denied, cluttered, and perception-limited environments, with future work focused on tighter fusion of state estimation across robots, semantics-informed planning and long-range marsupial autonomy.